\pdfoutput=1

\documentclass[11pt]{article}
\usepackage{EMNLP2022}
\usepackage{times}
\usepackage{latexsym}
\usepackage[T1]{fontenc}
\usepackage[utf8]{inputenc}
\usepackage{microtype}
\usepackage{inconsolata}
\usepackage{amsmath}
\usepackage{booktabs}
\usepackage{caption}
\usepackage{subcaption}
\usepackage{graphicx}
\usepackage{multirow}
\usepackage{soul}
\usepackage{multirow}
\usepackage[normalem]{ulem}
\useunder{\uline}{\ul}{}
\usepackage{tabularx}

\definecolor{col_table}{HTML}{249BFF}

\title{Weigh Your Own Words: Improving Hate Speech Counter Narrative Generation via Attention Regularization}

\author{Helena Bonaldi$^{1,2}$, Giuseppe Attanasio$^{3}$, Debora Nozza$^{3}$, Marco Guerini$^{2}$,  \\
  $^1$University of Trento, Trento, Italy \\
 $^2$Fondazione Bruno Kessler, Trento, Italy \\
 $^3$Bocconi University, Milan, Italy \\
\texttt{hbonaldi@fbk.eu}, \texttt{giuseppe.attanasio3@unibocconi.it},\\ \texttt{debora.nozza@unibocconi.it}, \texttt{guerini@fbk.eu}}

\begin{document}
\maketitle
\begin{abstract}

Recent computational approaches for combating online hate speech involve the automatic generation of counter narratives by adapting Pretrained Transformer-based Language Models (PLMs) with human-curated data. This process, however, can produce in-domain overfitting, resulting in models generating acceptable narratives \textit{only} for hatred similar to training data, with little portability to other targets or to real-world toxic language.
This paper introduces novel attention regularization methodologies to improve the generalization capabilities of PLMs for counter narratives generation. Overfitting to training-specific terms is then discouraged, resulting in more diverse and richer narratives. We experiment with two attention-based regularization techniques on a benchmark English dataset. Regularized models produce better counter narratives than state-of-the-art approaches in most cases, both in terms of automatic metrics and human evaluation, especially when hateful targets are not present in the training data. This work paves the way for better and more flexible counter-speech generation models, a task for which datasets are highly challenging to produce.

\noindent
\textit{\textbf{Warning}: this paper contains unobfuscated examples some readers may find offensive.}

\end{abstract}

\section{Introduction}

\begin{figure}[t!]
 \includegraphics[width=\columnwidth]{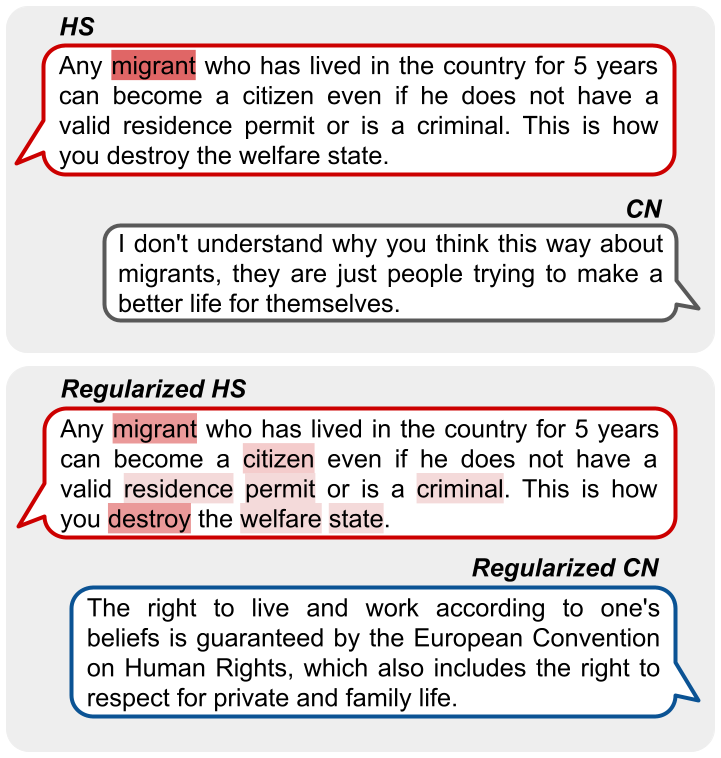}
 \caption{An example of CNs obtained with and without regularization: the highlighted terms show where the models focus their attention in the two cases.}
\label{fig:reg_example}
\end{figure}

Counter narratives (CNs) are an effective way to contrast hate speech as they are defined as ``communicative actions aimed at refuting hate speech through thoughtful and cogent reasons, and true and fact-bound arguments'' \citep{schieb2016governing}. In contrast to other widely used restrictive measures such as content removal and shadow-banning, CNs are based on the assumption that in order to combat hate speech, more speech is required. They are typically employed by Non-Governmental Organizations (NGOs) as an active strategy to intervene in online discussions where hateful content is present. Responding to micro and macro-aggressions with concrete action is critical because it can make such aggressions visible, disarm them, educate the perpetrators, and allow for external support \citep{sue2019disarming}. In particular, the key to the effectiveness of CNs is their specificity: they are more complex than a simple condemnation of profanity, and they include a variety of arguments \citep{tekirouglu2020generating}.

Still, the massive amount of hate that is constantly produced online necessitates the development of automatic CN generation techniques. Typically, this is done by fine-tuning a Pretrained Language Model (PLM) on human-curated data, such as GPT-2 \citep{radford2019language}. However, prior research has demonstrated that PLMs are susceptible to generating unspecific CNs that can technically work with any input but have questionable content and informativeness \citep{fanton2021human, tekiroglu-etal-2022-using}. We hypothesize that overfitting to training-specific terms is a possible cause of this behavior, as demonstrated for the task of hate speech detection \citep{attanasio-etal-2022-entropy}. In order to overcome this issue, this paper proposes two attention-based regularization approaches applied to a widely employed CN generation model on a benchmark English dataset \cite{fanton2021human}. The first strategy adapts to the generation scenario the Entropy-based Attention Regularization \citep[EAR; ][]{attanasio-etal-2022-entropy}, where the regularization term aims to maximize each token’s attention entropy. Then, we introduce a novel regularization technique called Kullback-Leibler Attention Regularization (KLAR), which makes the model pay particular attention to specific tokens connected to the stereotyped portrait of the minority targets. 

Figure \ref{fig:reg_example} shows an example of how attention regularization (EAR in this case) can induce the generation of richer CNs. The terms highlighted in the HS bubbles indicate where a CN generation model focuses its attention during generation.\footnote{See Figure \ref{fig:att_distrib_example} in Appendix \ref{attention_examples} for the full attention distribution on the input HS.}
The state-of-the-art CN generation model (top) poses the highest attention to the identity term ``migrant'', resulting in a vague response that would work with any hate speech targeting migrants. By redistributing the model's attention with the proposed attention regularization techniques (bottom), the focus shifts to include a broader context, resulting in the generation of a more specific and factually asserted CN. 

Moreover, we also assess the generalization abilities of the regularized models through a Leave One Target Out (LOTO) generation experiment. The results show that CNs generated with regularized models obtain better scores on standard automatic metrics, and they are considered more specific by human annotators, especially in LOTO settings. This is indicative of the robust settings of our proposed strategies. Code is available at \url{https://github.com/MilaNLProc/weigh-your-own-words}.

\section{Attention Distribution in CN generation} 
Attention is crucial in transformer language models. Loosely speaking, it regulates \textit{contextualization} of representations, i.e., how much of the context every token uses in its next-layer representation.
Such quantity is dictated by \textit{attention weights}: a higher weight will make a model focus more on a specific token, whereas zeroing out a token's attention will result in removing that contextual information.
Attention weights typically result from a training stage driven by data and a task objective, for example fine-tuning on a parallel corpus of hate speech and counter narratives. However, we hypothesize such a choice is sub-optimal and can lead to generic counter speech.

This section shortly reviews the role of attention in transformer LMs (\S\ref{app:attention_in_autoregressive}) and describes a preliminary study on attention distributions in CN generation models (\S\ref{sec:attention_cn_generation}).

\subsection{Attention in Autoregressive LMs}
\label{app:attention_in_autoregressive}

Autoregressive decoder-only LMs such as GPT2 \citep{radford2019language} are composed of \textit{L} transformer layers that process input token embeddings in cascade. Each layer transforms inputs using two sub-layers: a (multi-headed) attention layer and a subsequent point-wise multi-layer perceptron \citep[MLP; ][]{Vaswani2017AttentionIA}. While MLPs update token representations \textit{locally} \citep{geva2023dissecting}, attention regulates \textit{global contextualization}, i.e., \textit{which} part of the context, and \textit{how much} of it, each token will use.

We report the process undergoing attention sub-layers for an arbitrary token and layer.
Let $i$ be the positional index of the token under study and $C = \{c_0, c_1,...,c_{i}\}$ its left context.\footnote{As per standard formulation, the left context includes the token itself.}
The attention sub-layer builds a new token representation $s_i$ as
\begin{equation}
    s_i = \sum_{j=0}^i{a_{i,j} \cdot c_{j}}
\end{equation}
where $\mathbf{a}_i = \{a_{i,0}, a_{i,1}, ... , a_{i,i}\}$ is the set of attention weights from $i$ to every context token $c_j$.\footnote{For the sake of simplicity, we leave out some technical details such as the Query, Key, and Value projection matrices (Q, K, V). The reader can consider each $a_{i,j}$ the results of the scaled dot product between Q and K embeddings  and each $c_j$ the V projected embeddings.} Every attention set sums to one, i.e., $||a||_1 = 1$. Intuitively, a uniform attention distribution is equivalent to stronger contextualization, as more tokens contribute to $s_i$. Attention weights are hence a by-product of inference passes, i.e., they cannot be directly edited arbitrarily.

\subsection{Attention in CN Generation models}
\label{sec:attention_cn_generation}

\begin{table}[htbp]
\begin{center}\resizebox{\columnwidth}{!}{
\begin{tabular}{lcc}
\toprule
Measure       & \multicolumn{1}{c}{Normal} & \multicolumn{1}{c}{Relevant} \\ \midrule
mean attention         & 0.008                               & \textbf{0.013}                        \\
mean attention entropy & 3.387                               & \textbf{3.292}                        \\
mean std of attention  & 0.005                               & \textbf{0.008}       \\ \bottomrule                
\end{tabular}
}
\caption{Results of the analysis on the attention received by HS tokens.}
\label{tab:res_attention_HS}
\end{center}
\end{table}

We focus on the task of generating one counter narrative (CN) in reply to a hate speech (HS) input (see Figure \ref{fig:reg_example}). In this scenario, when generating a CN, a model uses the context given by the HS. 
In this section, we test if a PLM fine-tuned for CN generation poses a disproportionate quantity of attention on a specific set of terms, following previous work in hate speech detection \citep{attanasio-etal-2022-entropy}.
Hence, we analyze what happens during CN generation, from the perspective of (i) the attention received by HS tokens and (ii) the attention expressed by CN tokens.
The analysis was performed on a dataset of counter narratives that has been generated and human-evaluated in the work by \citet{tekiroglu-etal-2022-using}. This dataset includes 200 HS-CN pairs annotated with \textit{CN Specificity}, which measures how specific a CN is as a response to a particular HS. This allows us to determine if there is a correlation between attention and specificity, in particular if CNs with a disproportionate quantity of attention on specific terms are perceived as more vague.

In particular, we consider the set of relevant terms $R$ to study as the union of two sets: \textit{identity terms} and \textit{prejudice terms}. Identity terms are closely related to the identity of the targeted group (e.g., migrants), while prejudice terms are just part of stereotypical expressions related to that group (e.g., steal). Tokens that do not belong to the relevant terms were denoted as \textit{normal}. Appendix \ref{app:relevant_terms} reports the relevant terms list and its construction process. 

\paragraph{Attention HS tokens receive} First, we analyze the attention of HS tokens to inspect whether the relevant terms we identified are indeed relevant for the generative models.

Using the original \citet{tekiroglu-etal-2022-using} dataset splits, we fine-tuned GPT-2, and extracted the attention expressed by the model while generating the human-evaluated CNs (see Appendix \ref{finetune_hyp} for the fine-tuning details).
The process is performed as follows: the target text is split into tokens, each of which is appended incrementally for each forward pass. In this way, we extracted the set of attention weights expressed at each generation step -- forward pass -- towards all the preceding tokens (HS included, since we employ autoregressive models)\footnote{This incremental procedure is independent from which next token was actually selected, i.e. it does not matter which decoding mechanism was employed for generating the target text. In fact, the attention weights are expressed during the forward pass, before next-token selection, thus
only the fine-tuned model and the input text are needed to compute the attention.}.  

Our analysis focused on three different values: (i) \textit{mean attention}, which measures the importance of each token category for the model during generation; (ii) \textit{mean attention entropy}; and (iii) \textit{mean standard deviation of attention}, which considers whether the quantity of received attention is consistent across decoding steps or subject to peaks.
The mean is performed by averaging scores across all layers and heads. 
We used a t-test for independent samples with a two-sided alternative hypothesis to test if the results for the relevant terms differed significantly from those for the normal terms\footnote{This was possible since the variance of both the mean attention and of the mean attention entropy was similar across the tokens categories.}.

Table \ref{tab:res_attention_HS} shows that relevant terms in the HS receive significantly higher mean attention than normal tokens, thus showing to be particularly important for the model during CN generation. Moreover, they have a significantly lower mean attention entropy of received attention than normal tokens: therefore, these terms are important only in specific decoding steps.

\begin{figure}[t!]
 \includegraphics[width=\columnwidth]{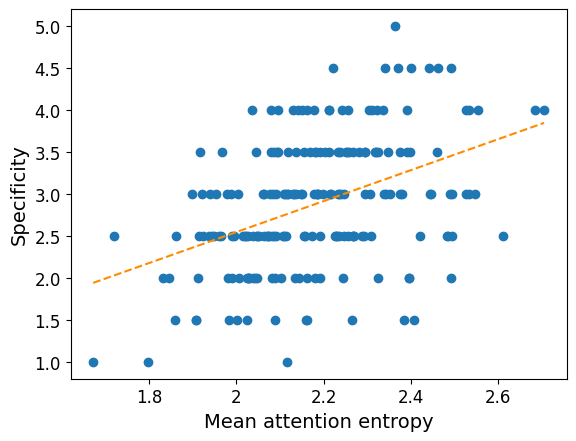}
 \caption{The correlation between the mean attention entropy and the specificity of CN generated with GPT-2.}
\label{fig:correlation_entropy_spec}
\end{figure}

\paragraph{Attention CN tokens express} The second focus of our analysis is the attention expressed by CN tokens towards previous tokens during generation. In particular, we test whether there is a correlation between the distribution of the CN attention (i.e., the CN attention entropy) and the human-evaluated \textit{CN specificity}. Following the work by \citet{attanasio-etal-2022-entropy} we hypothesize that a low attention entropy is related to a vaguer generation, and thus to a lower quality. While a higher attention entropy, and consequently a more uniform attention distribution is associated with more specific generations.

We first computed the \textit{mean attention entropy} value for each CN by averaging the attention entropy of each token that comprised it.
 Then, we calculated Spearman's correlation between the computed CNs attention entropy and the specificity values. Figure \ref{fig:correlation_entropy_spec} shows a direct correlation of 0.42 between attention entropy and specificity (with exact $p$-value < 0.01\footnote{We employed a permutation test to calculate the p-value, since the dimension of the two compared sets were different.}). This confirms our hypothesis: \textbf{the higher the attention entropy, the higher the specificity of the generated CN}. This motivated the need for model attention regularization to consider a wider context, as presented in the following section.

\section{Attention Regularization for Counter Narrative Generation}
\label{sec:}

This section presents the two proposed regularization solutions  to steer models toward the generation of more specific counter narratives.

\paragraph{Entropy-based Attention Regularization}

Originally introduced in \citet{attanasio-etal-2022-entropy}, Entropy-based Attention Regularization (EAR) adds a penalization term to encoder language models as a function of attention weight distributions. Roughly, EAR penalizes the model whenever a token's self-attention weights have a low-entropy distribution. The authors use then a loss
$$
    \mathcal{L} = \mathcal{L}_C - \alpha \cdot \frac{1}{|L||C|} \sum_{l \in L}{\sum_{i \in C}{H_i^l}} 
$$
where $\mathcal{L}_C$ is the standard cross entropy loss, $\alpha$ is a scalar to set the regularization strength\footnote{The higher the $\alpha$ value, the more uniform is the attention distribution that the model is forced to have.}, $H_i^l$ is the attention entropy for the \textit{i}-th token and \textit{l}-th layer, and \textit{C} the considered context. Intuitively, by teaching the model to use higher entropy, EAR forces a stronger contextualization of token representations. The authors prove that EAR reduces lexical overfitting on group-specific identity terms in hate speech detection. 

To apply EAR to decoder language models and text generation, we consider \textit{C} to be each token's left context and $\mathcal{L}_C$ the cross-entropy language modeling loss over our vocabulary.

\paragraph{Kullback-Leibler Attention Regularization}

\begin{figure}[!t]
    \centering
    \includegraphics[width=\linewidth]{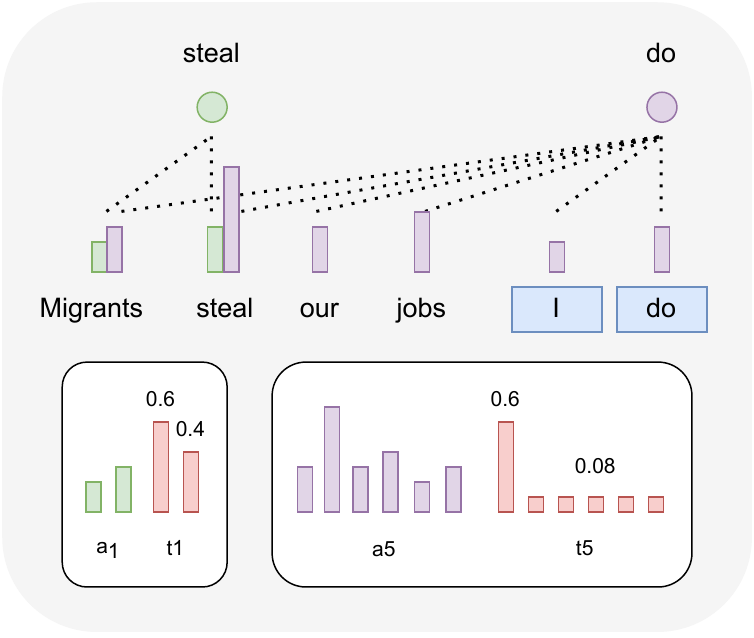}
    \caption{KLAR for next-layer representations of the tokens ``steal'' (green bars) and ``do'' (purple), with $R = \{\text{Migrants}\}$ and $\lambda = 0.6$. Blue boxes are generated tokens and dotted lines signal existing attention weights. Bottom boxes represent $KL(\cdot)$ operations between real attention distributions ($\mathbf{a_1, a_5}$) and target ($\mathbf{t_1, t_5}$, red) attention distributions.}
    \label{fig:klar}
\end{figure}

EAR regularizes every token with a strong assumption: having a uniform attention distribution will let the model avoid overfitting to specific words, regardless of their context.
However, we hypothesize that in addition to benefiting from more contextualization, counter narrative generation needs to ``prioritize'' specific relevant words. Ideally, such words will provide context and guidance for a more targeted and richer counter narrative.

Thus, we reformulate attention regularization to account for word prioritization by introducing Kullback-Leibler Attention Regularization (KLAR). KLAR is a training-time regularization approach that \textbf{steers models to use specific attention distributions}.
KLAR adds an auxiliary loss term to the standard language modeling loss. We compute the regularization loss upon attention weights as follows.\footnote{We introduce KLAR in decoder-only models where attention is allowed to the left context only. KLAR is extensible to sequence-to-sequence models with minimal edits.}

Let $\mathbf{a_{i}} = \{a_{i,0}, a_{i,1}, ... a_{i,i}\}$ be the set of attention weights from $i$ to every left-token $c_j$. As in \citet{attanasio-etal-2022-entropy}, we average weights over attention heads and apply the softmax to the result to restore a probability distribution.
Then, let $R \subseteq C$ be the subset of \textit{relevant} words and $N = C \setminus R$ its complement set of non-relevant ones, such that $R \cap N = \emptyset$ and $R \cup N = C$. 

Next, we introduce the notion of \textit{target attention distribution} $\mathbf{t}_i = \{t_{i,0},t_{i,1},...,t_{i,i}\}$. This distribution reflects our expectation of what a \textit{gold} attention distribution should look like. In KLAR, we use $R$ and $N$ to derive $\mathbf{t}_i$ as follows:
\[ t_{i,j} =
  \begin{cases}
    \lambda / |R|      & \quad \text{if } c_j \in R \\
    (1 - \lambda) / |N|  & \quad \text{if } c_j \in N
  \end{cases}
\]
where $\lambda$ is an ``attention share'' we assign to relevant tokens. Notably, tokens within a set, either $R$ or $N$, share the same target attention weight. The choice of $R$, $N$, and $\lambda$ is part of our experimental setup. Figure~\ref{fig:klar} shows KLAR graphically on the previous example.

Finally, we define the KLAR regularization term as
$$
    \mathcal{L}_{i,KLAR} = \frac{1}{|L|}\sum_{l \in L}{KL(\mathbf{a}_i^l||\mathbf{t}_i^l)}
$$
where $L$ is the set of transformer layers, $a_i^l$ and $t_i^l$ are the real and target attention distributions at each layer, respectively, and $KL(\cdot)$ is the Kullback-Leibler divergence function. Note that the function is fully differentiable. We do not change $t_i^l$ across layers.

Autoregressive decoder-only language models compute one loss contribution per input token. We follow a similar approach and compute $\mathcal{L}_{KLAR}$ for every token and sum it to the task loss. Hence, the final loss at position $i$ is:
$$
    \mathcal{L}_i = \mathcal{L}_{LM} + \alpha \cdot \mathcal{L}_{i,KLAR}
$$
where $\mathcal{L}_{LM}$ is the standard language modeling loss computed as cross-entropy over the token vocabulary. As for EAR, $\alpha$ controls the regularization strength.

\section{Experimental setup}

\subsection{Models and decoding mechanisms} 

We use GPT-2 \citep{radford2019language}, a decoder-only autoregressive PLM, for two reasons. First, it has a masked self-attention mechanism that is only directed to the left context. Other Transformer-based PLMs, such as encoder-decoder architectures, include an additional encoder-decoder attention mechanism, which would complicate our analyses even more. Second, because it is one of the most widely used PLMs for the CN generation task \cite{fanton2021human, tekiroglu-etal-2022-using}.

We test several decoding mechanisms: beam search \citep{li-etal-2016-deep, wiseman-etal-2017-challenges}, top-$k$ \cite{fan-etal-2018-hierarchical}, top-$p$ \citep{holtzman2019curious}, the combination of top-$k$ and top-$p$, and contrastive search \citep{su2022a}. Driven by the results obtained from preliminary experiments on the validation set, we chose to employ beam search (\textsc{bs}) and contrastive search (\textsc{con}). The selection of the decoding mechanisms and regularizations hyperparameters is described in Appendix \ref{hyp_selection}.

\subsection{Datasets}
We use the benchmark counter speech dataset Multi-Target CONAN (MTCONAN) \cite{fanton2021human}, which is the most varied and high quality available dataset. The dataset consists of 5003 hate speech (HS) and counter narratives (CNs) pairs covering several targets of hate, including Disabled, Jews, LGBT+, Migrants, Muslims, People Of Color (POC), and Women. The data were collected via a human-in-the-loop strategy, starting from a seed dataset niche-sourced from 20 experts from two different NGOs.

\subsection{Evaluation Metrics} \label{eval_metrics}
We evaluate the generations produced by the automatic CN generation models using both standard automatic metrics and a comprehensive human evaluation study.

\paragraph{Diversity} We measure the lexical diversity of the generated data with the Repetition Rate (\textsc{RR}), which quantifies the average ratios of non-singleton \textit{n}-grams in the overall generation \citep{cettolo2014repetition, bertoldi2013cache}. We use it to determine which strategy is more effective in providing diverse generations. 

\paragraph{Overlap with gold reference} We consider the similarity to the gold CNs in MTCONAN  as a proxy of the quality of our generations. We use several similarity metrics: BLEU-1, BLEU-3, BLEU-4 \citep[\textsc{B1}, \textsc{B3}, \textsc{B4}; ][]{papineni2002bleu} and ROUGE-L \citep[\textsc{RL}; ][]{lin2004rouge} , which are based on n-gram similarity. This allows us to test both lexical and semantic similarities between the generations and the gold references.

\paragraph{Human evaluation} Automatic metrics provide a limited overview of the quality of open-domain generations \cite{belz2006comparing, novikova2017we}. In particular, the comparison with the gold references does not allow for an understanding of the overall quality of the generated CNs. Thus, we conduct a human evaluation study with 7 annotators. Before starting the evaluation, we explained in detail the task to the annotators and made them read several examples which were manually created by expert NGO operators to make them understand how proper CNs are written. We also organized meetings to discuss with them, in order to allow possible problems and stress to emerge, following a mitigation procedure similar to the one proposed by \citet{vidgen-etal-2019-challenges}.

We ask annotators to measure the following desired characteristics in the generated CNs: (i) \textit{Suitableness} (\textsc{SU}) quantifies how much the CN follows a specific style and guidelines, which are an adaptation of those used for the European project and campaign \textit{Get The Trolls Out!}.\footnote{\url{https://getthetrollsout.org/stoppinghate}} Some examples of these guidelines are ``Don't be abusive'', ``Don't get personal'' and ``Think about your tone''. (ii) \textit{Specificity} (\textsc{SP}) indicates how specific a CN is in responding to the HS. For example, a very vague CN such as \textit{Do you have any evidence for this?} would obtain a very low specificity score. These two measures are rated with a 1-5 Likert scale, following \citet{chung2020italian}. Moreover, the annotators are asked to rank (\textsc{avg rank}) the model generations according to their overall quality.

\subsection{Experimental configurations}

We evaluated the proposed regularization approaches on two data configurations: (i) \textit{in-target} which follows the conventional training and test splits and (ii) \textit{out-of-target} which creates splits in which one target is absent from the training data but present in the test data.

\paragraph{In-target CN generation}
First, we test our proposed regularization techniques on the traditional fine-tuning task, where we train and test GPT-2 on the original MTCONAN splits, thus with a 8:1:1 proportion. 
Each split contains all targets, in a proportion reflecting the overall distribution of targets in the dataset. The fine-tuning details are reported in Appendix \ref{finetune_hyp}. During the generation, only the HS is given as input, and we generate one CN in response to each HS in our test set (thus 500 CNs). Human evaluation is then performed on a subset of 420 examples, since it is more resource-intensive.

\paragraph{Leave One Target Out CN generation}
In order to test how the proposed regularization techniques improve the model's ability to generalize to unseen targets, we put into practice an out-of-target experiment. In particular, we replicate the setup of the Leave One Target Out (LOTO) experiment performed by \citet{tekiroglu-etal-2022-using}. This experiment consists in testing a model over the data pertaining to a target that is not present in the training set. First, we select the targets which are most prominently present in the MTCONAN dataset, in order to have a sufficiently large test set for each configuration. Thus, we sampled from MTCONAN 600 examples for each of the following targets: Muslims, Migrants, Women, LGBT+, and Jews\footnote{The data corresponding to POC, Disabled, and the \textit{other} target were also included in the dataset and used only for training. We only excluded examples containing a reference to multiple targets)}. The resulting dataset is composed of 3729 HS and CN pairs. Next, we fine-tuned 5 models by using as training data all the examples except those referred to one of the 5 targets mentioned above.\footnote{We used the same hyperparameters employed in the previous experiment.} The data covering the left-out target constitute the test data for the generation. For this experiment, human evaluation was performed on 380 examples.

\section{CN generation results}
This section discusses the outcomes of both experiments, as well as the circumstances under which regularization of the CN generation task is appropriate. Some examples of the generations are shown in Table \ref{tab:gen_ex} in Appendix \ref{gen_examples}.

\begin{table*}[t!]
    \begin{center}\resizebox{\textwidth}{!}{
    \begin{tabular}{ll|c|rrrc|rrc}
    \toprule
    Deco.       & \multicolumn{1}{c}{Reg. }         & \multicolumn{1}{c}{\textbf{} }            & \multicolumn{4}{c}{Overlap}           & \multicolumn{3}{c}{Human evaluation}                                                             \\ \midrule
                         & \multicolumn{1}{c|}{\textbf{}} & \multicolumn{1}{c|}{\textsc{RR}}   & \multicolumn{1}{c}{\textsc{RL}} & \multicolumn{1}{c}{\textsc{B1}} & \multicolumn{1}{c}{\textsc{B3}} & \multicolumn{1}{c|}{\textsc{B4}}     & \multicolumn{1}{c}{\textsc{SU}} & \multicolumn{1}{c}{\textsc{SP}} & \multicolumn{1}{c}{\textsc{avg rank}} \\ \midrule 
    \multirow{3}{*}{\textsc{con}} 
                        & No-Reg            & \textbf{11.810}               & 0.158                           & 0.159                           & 0.018                           & 0.009                                 & 3.462                           & \textbf{2.621}                  & 2.029                                 \\ 
                         & KLAR              & 12.862                  & \textbf{0.161}                  & \textbf{0.160}                  & \textbf{0.022}                  & \textbf{0.011}                       & \textbf{3.562}                  & 2.471                           & \textbf{1.967}                        \\
                         & EAR               & 13.091              & 0.157                           & 0.157                           & 0.020                           & 0.009                                         & 3.546                           & 2.454                           & 2.000                                 \\ \midrule
    \multirow{3}{*}{\textsc{bs}}  
                        & No-Reg         & \textbf{16.560}                 & \textbf{0.156}                  & \textbf{0.158}                  & 0.023                           & 0.009                         & 3.683                           & 2.600                           & 1.983                                 \\
                         & KLAR            & 21.511                     & \textbf{0.156}                  & 0.150                           & 0.024                           & \textbf{0.013}                                & 3.594                           & 2.522                           & 2.017                                 \\
                         & EAR                 & 17.853                  & 0.154                           & 0.154                           & \textbf{0.027}                  & \textbf{0.013}                             & \textbf{3.694}                  & \textbf{2.617}                  & \textbf{1.967}      \\ \bottomrule                 
    \end{tabular}
    }
    \caption{Results of the In-Target generation experiment.
    }
    \label{tab:res_indom}
    \end{center}
    \end{table*}

\paragraph{In-target generation}

Table \ref{tab:res_indom} shows the results in terms of the presented metrics for the model without regularization (No-Reg), KLAR and EAR, with both contrastive search and beam search decoding.

Regarding the contrastive search setup, KLAR achieves the highest degree of overlap with gold data. This comes at the expense of the \textsc{RR} and specificity, where it performs second best. In general, the highest suitableness and average rank indicate an overall better quality of CNs generated with KLAR. 

With beam search, the results obtained by the various configurations on the overlap scores are similar. This may be due to the stricter probability constraint imposed by the beam search decoding on the generation: the deterministic sampling is most likely causing similarly good word choices despite the performed regularization. However, a clearer pattern emerges if we consider the human evaluation scores: for all the evaluated dimensions, EAR obtains the best scores. This again comes at the expense of a marginally higher \textsc{RR}.

\paragraph{Leave One Target Out generation}

\begin{table*}[t!]
    \begin{center}\resizebox{\textwidth}{!}{
    \begin{tabular}{ll|c|rrrc|rrc}
    \toprule
    Deco.       & \multicolumn{1}{l}{Reg. }         & \multicolumn{1}{c}{\textbf{} }            & \multicolumn{4}{c}{Overlap}           & \multicolumn{3}{c}{Human evaluation}                                                             \\ \midrule
                         & \multicolumn{1}{c|}{\textbf{}} & \multicolumn{1}{c|}{\textsc{RR}}   & \multicolumn{1}{c}{\textsc{RL}} & \multicolumn{1}{c}{\textsc{B1}} & \multicolumn{1}{c}{\textsc{B3}} & \multicolumn{1}{c|}{\textsc{B4}}     & \multicolumn{1}{c}{\textsc{SU}} & \multicolumn{1}{c}{\textsc{SP}} & \multicolumn{1}{c}{\textsc{avg rank}} \\ \midrule 
    \multirow{3}{*}{\textsc{con}}            
                                    & No-Reg        & \textbf{13.467}                      & 0.145                           & 0.141                           & 0.012                           & 0.004                                                                  & 3.450                           & 2.183                           & 2.050                                 \\
                                      & KLAR           & 14.073                             & \textbf{0.146}                  & \textbf{0.143}                  & \textbf{0.014}                  & \textbf{0.006}                                                            & 3.471                           & 2.254                           & 2.017                                 \\
                                      & EAR               & 14.075                         & 0.145                           & 0.140                           & \textbf{0.014}                  & 0.005                                                            & \textbf{3.638}                  & \textbf{2.262}                  & \textbf{1.933}                        \\ \midrule
    \multirow{3}{*}{\textsc{bs}}              
                             & No-Reg             & \textbf{20.044}                   & 0.139                           & 0.137                           & 0.017                           & 0.007                                                                        & 3.521                           & 2.100                           & 2.021                                 \\
                                      & KLAR         & 21.426                          & \textbf{0.147}                  & \textbf{0.148}                  & \textbf{0.019}                  & \textbf{0.009}                                                              & \textbf{3.700}                  & 2.129                           & \textbf{1.936}                        \\
                                      & EAR            & 22.159                      & 0.138                           & 0.131                           & 0.017                           & \textbf{0.009}                                                                         & 3.514                           & \textbf{2.286}                  & 1.964          \\ \bottomrule                      
    \end{tabular}
    }
    \caption{Results of the LOTO generation experiment.
    }
    \label{tab:res_loto}
    \end{center}
    \end{table*}

In Table \ref{tab:res_loto}, the average scores obtained by the 5 LOTO models we tested are shown.
Regularization seems to be particularly beneficial in the LOTO setup, according to both automatic and human evaluation. EAR is the configuration achieving the highest specificity for both decoding strategies. In addition, KLAR achieves the highest overlap scores, demonstrating its ability to generate CNs that are more lexically similar to human-written ones. When considering the best suitableness and average rank, EAR performs better with contrastive search, while KLAR with beam search. With both decoding mechanisms, choosing not to regularize generates CNs with the lowest specificity and overall quality (worst average rank).

\paragraph{General Discussion}
Across the two performed experiments, some common phenomena can be noticed. Overall, \textbf{regularization is preferable in terms of both automatic and human evaluation}. KLAR is the best performing model on automatic metrics, while EAR allows obtaining higher human-evaluated specificity in most cases. The results are particularly clear-cut for the LOTO setup, showing the robustness of our proposed techniques in generalizing to unseen targets.

In both experiments, KLAR obtains the highest overlap scores with gold CNs. This indicates that focusing on relevant terms produces words that are more lexically similar to human-written data.

In addition, the human evaluation measures favor EAR, especially in terms of specificity. This is particularly useful for the CNs generated with beam search. Even if such CNs tend to be slightly more repetitive than those obtained with contrastive search (as shown by the higher average \textsc{RR}) they are also safer, as shown by \citet{tekiroglu-etal-2022-using}. In this safer configuration, EAR allows reaching a higher specificity without a big impact on the overall quality (average rank).

The best suitableness and average rank are always reached by a regularized model: KLAR for the in-target experiment with contrastive search and LOTO with beam search, and EAR for the in-target experiment with beam search and for LOTO with contrastive search. One possible interpretation is that the model requires a stronger regularization in the most extreme cases for these dimensions. In particular, when the generation is the most constrained (i.e. in the in-target experiment with beam search) and the freest/most undecided (i.e. in LOTO with contrastive search), imposing a uniform attention distribution with EAR seems to be preferable. On the other hand, in those that we can consider as ``middle cases'' (i.e. in-target experiment with contrastive search and LOTO with beam search), where either the model has clearer word choices because of the in-domain training or because of a deterministic decoding, regularizing on specific terms with KLAR is sufficient.

In general, \textbf{regularizing shows to be effective in improving the generalization capabilities of the model when faced with unseen targets}, confirming the results obtained by \citet{attanasio-etal-2022-entropy} also for the generation task. This is shown by the clearer pattern of better scores obtained in the LOTO experiment, and by a higher difference between the mean rank of the best scoring regularization technique and No-Reg\footnote{In LOTO the average difference between the mean rank of the best scoring regularization technique and No-Reg is of 0.1025, as opposed to 0.056 in the first experiment}.

\section{Related Work}

\paragraph{Counter Narrative Generation}
The effectiveness of employing CNs to fight online hate has brought a growing interest towards automating the CNs production. The most widely employed strategy consists in fine-tuning a PLM using ad hoc created data, in either a single \citep{qian-etal-2019-benchmark, fanton2021human, zhu2021generate, tekiroglu-etal-2022-using} or multi-turn setting \citep{de2021toxicbot, bonaldi-etal-2022-human}.  Some of these works have focused on specific desired characteristics of the generated CNs: \citet{chung-etal-2021-towards} worked on knowledge-grounded generation, \citet{de2021toxicbot} accounted for personality and \citet{ijcai2022p716} controlled for politeness, detoxification and emotion. Differently from this trend, \citet{chung2020italian} experimented with zero and few-shots learning configurations for generating Italian CNs. Finally, \citet{ashida2022towards} used few-shots and one-shot prompting of large models (GPT-Neo, -2, -3) to generate both CNs and microinterventions. 

\paragraph{Language Models Regularization}
Recent research has investigated the use of regularization losses to improve language models. Most generally, \citet{su2022a} introduce an auxiliary contrastive learning-based term to improve model pretraining. In the context of policy learning, \citet{ouyang2022training} use multiple auxiliary losses to retain good language modeling capabilities while optimizing for rewards on human preferences. Other relevant works focus on regularization for robust fine-tuning \citep[\textit{inter alia}]{aghajanyan2021better,jiang-etal-2020-smart,dong2021should}. Our approach is closer to the latter, although we require no additional inference pass or noise injection.

\section{Conclusion}
In this work, we proposed two novel regularization techniques for the task of counter narrative generation. Our aim is to avoid PLM's tendency to overfit to specific terms, thereby producing vague and thus ineffective counter narratives, as evidenced by our preliminary analysis. To the best of our knowledge, this is the first time that based attention regularization is employed for improving CN generation. We also make a novel contribution by introducing the Kullback-Leibler Attention Regularization (KLAR). Results show that human annotators tend to consider CNs generated with regularized models more specific and suitable in most cases, especially in out-of-target settings. CNs obtained by regularized models also have better scores on standard automatic metrics. This work can be considered as an initial step towards a more adaptable generation of CNs.

\section*{Limitations}
Most available data for CN generation are in English, thus, in this work, we focus solely on this language. Moreover, even if the quality of the MTCONAN dataset is high and the generations have been checked by expert NGOs operators, its variety is still limited. Thus, the improvements obtained with our regularization are also limited by this aspect.

In this work, we make use of lists of manually identified relevant terms: however, this reduces the effectiveness of our strategies applicability. For this reason, in future work we plan to investigate how to automatically extract such terms. We also mainly focused on negative prejudice terms, but another interesting development of this work could analyze the usefulness of more positive terms typically employed in CNs. Finally, despite the improvements shown by our proposed regularization techniques over the traditional fine-tuning, regularizing brings a higher \textsc{RR}: our next objective would thus be to obtain higher specificity values without impacting the \textsc{RR}.

\section*{Ethics Statement}

The main aim of this paper is to develop novel methods and techniques to aid the manual work of NGOs operators. Thus, all the systems we develop are not intended for being used in the wild but in a human-machine collaboration setting, allowing the human to check and possibly post-edit the machine-generated CNs. 

The focus of our work is to develop effective CNs: for this reason, and to avoid possible misuse of our models, the data we employed for training the HS are simple and stereotyped. Moreover, we will not distribute the generated data since their content has not been thoroughly checked by human annotators. As regards the annotation process, since we are aware that long exposure to hateful data might be stressful and hurtful, we put in practice a mitigation procedure similar to that proposed by \citet{vidgen2019challenges}, in order to preserve the mental health of the annotators.  

\bibliography{anthology,custom}
\bibliographystyle{acl_natbib}

\clearpage
\appendix

\section{Appendix}
\label{sec:appendix}

\subsection{Generation examples} \label{gen_examples}
In Table \ref{tab:gen_ex} some examples of the generations obtained with and without the proposed regularization techniques are shown. The first row example is the same as that in Figure \ref{fig:reg_example}. In general, we can notice how No-Reg CNs tend to negate the HS (1, 3), vaguely indicate that we should take action (4) or completely fail to contrast the HS. KLAR and EAR instead, show to be more complex and articulated.

\begin{table*}[htbp]
\centering
\begin{tabularx}{\textwidth}{lXXXX}
\toprule
\multicolumn{1}{c}{}
&
\multicolumn{1}{c}{HS }                                                                                                                                                                             & \multicolumn{1}{c}{No-Reg}                                                                                         & \multicolumn{1}{c}{KLAR}                                                                                                                                                                     & \multicolumn{1}{c}{EAR}                                                                                                                                                       \\ \midrule
(1) &
Any migrant who has lived in the country for 5 years can become a citizen even if he does not have a valid residence permit or is a criminal. This is how you destroy the welfare state. & I don't understand why you think this way about migrants, they are just people trying to make a better life for themselves. & I don't see how it would be possible to "destroy the welfare state" by forcing people to apply for citizenship.                                                                   & The right to live and work according to one's beliefs is guaranteed by the European Convention on Human Rights, which also includes the right to respect for private and family life.  \\ \midrule
(2) &
Blacks always complain about racism saying we still treat them badly. But they are always trying to steal our jobs.                                                                      & I think it's important to remember that racism is not a problem in the UK.                                                  & The problem with racism is that it can lead to discrimination and violence, and it is not the fault of the person who has the problem. It is the fault of the society in general. & You can't blame people of colour for their own discrimination. It is a human rights issue, and it is a human right to be able to work.                                                 \\ \midrule
(3) &
Muslims have no place in civilised Europe.                                                                                                                                               & The idea that a religion is incompatible with civilised society is just nonsense.                                           & I don't see how you can say that about a religion that is so diverse and diverse in its beliefs and practices, and that has so much to offer to humanity.                         & I don't see how you can say that. The vast majority of European countries have a strong Christian heritage, and this has nothing to say about the presence of a minority in a country. \\ \midrule
(4) &
Migrants steal our jobs.                                                                                                                                                                 & The problem is that we have a system of immigration that is not working for us. We need to change it.                       & The fact that some migrants steal our jobs does not mean that they are stealing our culture.                                                                                      & The problem is that we are not able to cope with the influx of people. We need to integrate people into our society, but this is not possible if we do not respect their rights.      \\ \bottomrule
\end{tabularx}
\caption{Some examples of generation with No-Reg, EAR and KLAR.}
\label{tab:gen_ex}
\end{table*}

\subsection{Relevant terms}
\label{app:relevant_terms}
Table \ref{tab:rel_terms} shows the full list of relevant terms (i.e., the combination of identity and prejudice terms) that we employed for each target. Note that identity terms can also comprise offensive terms used to identify a particular group.
In order to extract the lists of identity and prejudice terms, we first take the 50 most frequent lemmas for each subset corresponding to a hate target in MTCONAN \citet{fanton2021human}. Then, 3 annotators manually annotated whether each term belonged to one of the two mentioned sets. Any inconsistency was then resolved with an internal discussion.

\subsection{Attention distribution examples} \label{attention_examples}
\begin{figure*}[t!]
\centering
   \begin{subfigure}{1\textwidth}
   \includegraphics[width=1\textwidth]{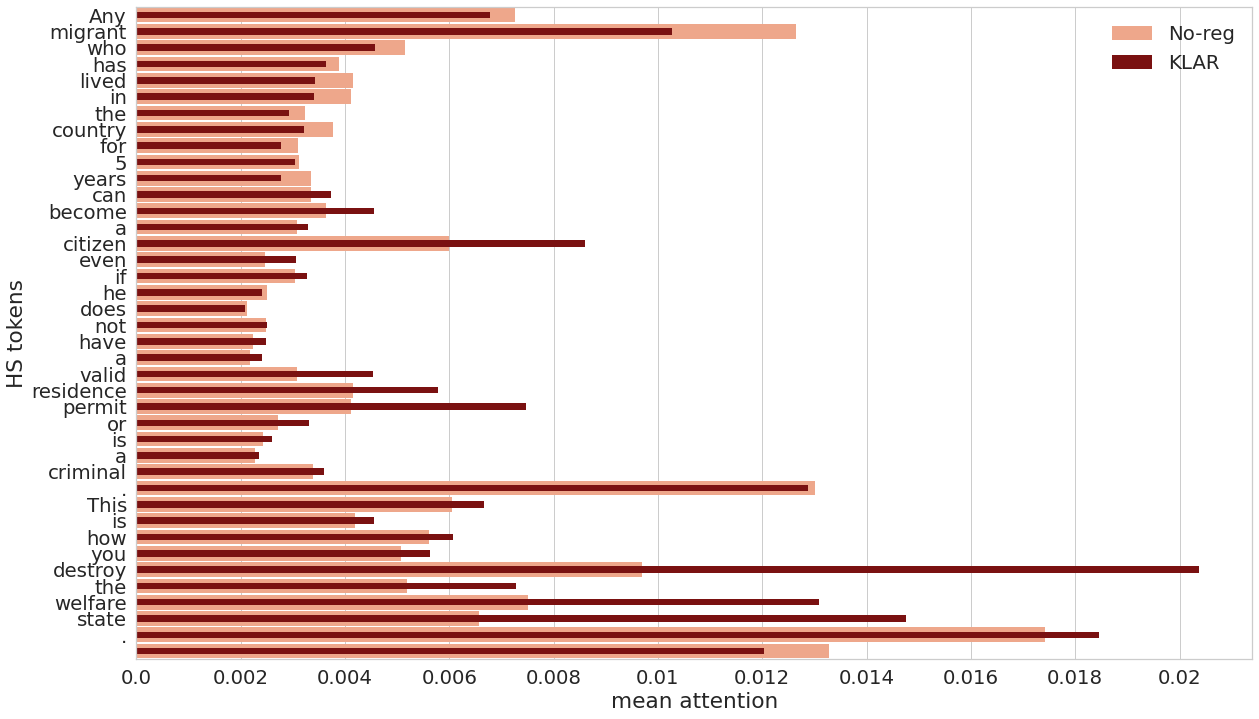}
\end{subfigure}

\begin{subfigure}{1\textwidth}
   \includegraphics[width=1\textwidth]{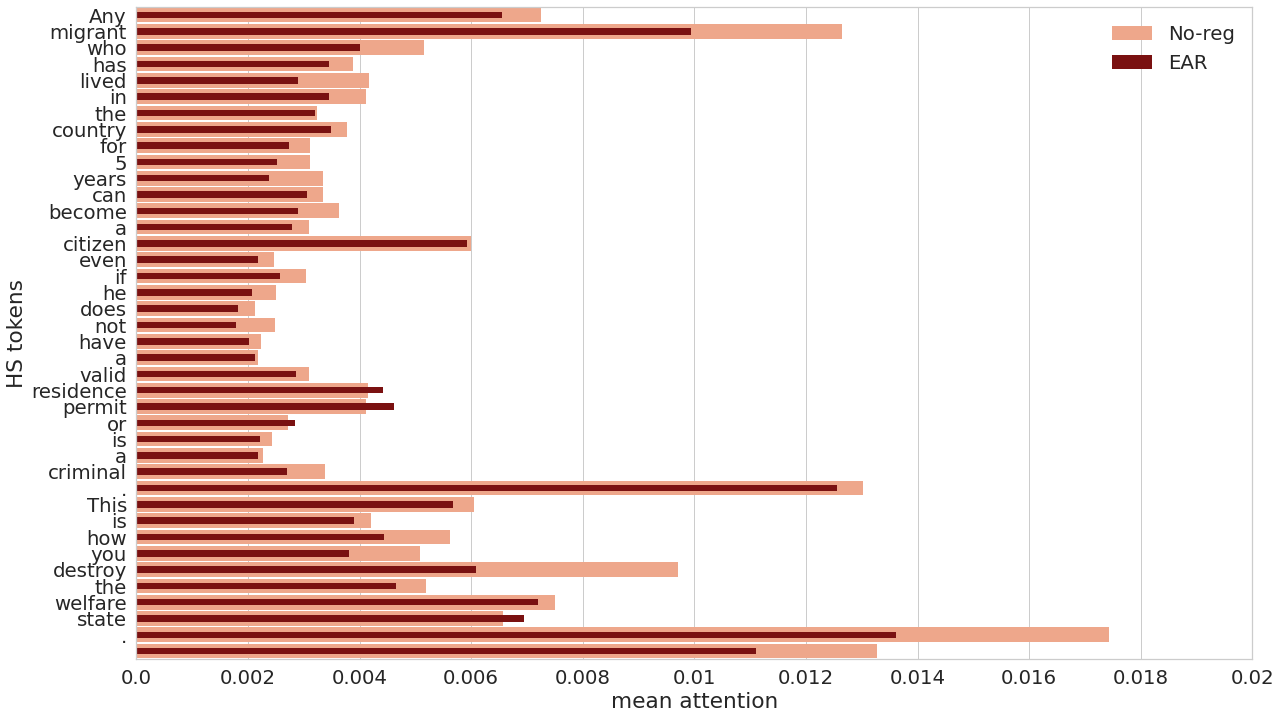}
\end{subfigure}
\caption{The attention distribution over the same HS of the No-Reg model, compared to KLAR and EAR.}
\label{fig:att_distrib_example}
\end{figure*}

In Figure \ref{fig:att_distrib_example} we show the distribution of the average attention of non-regularized GPT-2 while generating the first example in Table \ref{tab:gen_ex} and Figure \ref{fig:reg_example}, compared to the distribution of KLAR (a) and EAR (b). Note that for each model we show the average attention expressed while generating the respective CNs, shown in the first row of Table \ref{tab:gen_ex}.

If we do not consider punctuation, it is possible to see how the highest attention for the non-reg model is focused on the identity term ``migrant''. On the other hand, with EAR and KLAR the distribution changes fundamentally. With KLAR there is a higher attention concentrated on specific terms, such as \textit{destroy}, \textit{welfare}, \textit{state}, \textit{citizen}. With EAR, instead, attention is more uniformly lower, and slightly higher for a few terms only (\textit{residence}, \textit{permit}, \textit{state}). 

\begin{table*}[htbp]
\centering
\begin{tabularx}{\textwidth}{lXX}
\toprule
\multicolumn{1}{l}{Target} & \multicolumn{1}{c}{Identity terms}                                 & \multicolumn{1}{c}{Prejudice terms}                                                                                                                                                        \\ \midrule
Disabled                            & disabled,  disability,  autistic,  blind,  retard,  downies                 & child, syndrome, education, isolated, school, learning, mentally, difficulty, normal, mental, work, discrimination, burden, different, physical, excluded, physically, development, health, ability \\ \midrule
Jews                                & jew, jewish, holocaust, judaism                                             & religion, control, different, government, belief, nazi, race, hollywood, religious, law, culture, rule, power, attack                                                                               \\ \midrule
LGBT+                               & gay,  homosexual,  homosexuality,  lgbt,  faggot,  gender,  lesbian,  lgbti & sexual, child, sex, orientation, love, sick, isolated, disease, different, pervert, burden, preference, family, heterosexual, blood                                                                 \\ \midrule
Migrants                            & migrant, immigrant, refugee, immigration, foreigner, migration              & job, work, steal, culture, crime, tax, rape, citizen, burden, economy, criminal, war                                                                                                                \\ \midrule
Muslims                             & muslim, islam                                                               & religion, violent, incompatible, western, terrorist, culture, violence, peaceful, ideology, civilised, different, belief, peace, law, faith, woman, rape, death, child, cult                        \\ \midrule
POC                                 & black, nigga, african, afro-american, nigger, negro                         & white, color, race, skin, different, criminal, racism, crime, inferior, violence, violent, slavery, racist, genetically, subhuman               \\ \bottomrule                                                   
\end{tabularx}
\caption{The identity and prejudice terms we employed.}
\label{tab:rel_terms}
\end{table*}

\subsection{Fine-tuning setup} \label{finetune_hyp}
We employed a GPT-2 medium model, fine-tuned with the following hyper-parameters: training batch size $= 8$, evaluation batch size $ = 4$, number of training epochs $=3$, warmup ratio $=0.1$ and learning rate $=5e-05$. Similarly to other previous work \cite{fanton2021human, bonaldi-etal-2022-human}, in order to train GPT-2 for the task of CN generation, we employed special tokens. In particular, the training data were represented as: \\
\texttt{<hatespeech>} \ HS \ \texttt{<counternarrative>} \ CN \ \texttt{<|endoftext|>} \\
This data representation was replicated when giving the HS in input to the model for generation.

\subsection{Hyperparameter tuning for No-Reg, KLAR, EAR} \label{hyp_selection}

\begin{table*}[htbp]
\centering
\resizebox{\textwidth}{!}{
\begin{tabular}{llcccrrrrrl}
\toprule
Reg.           & Deco. & k & $\alpha$ & share & \multicolumn{1}{c}{\textsc{RR}} & \multicolumn{1}{c}{\textsc{RL}} & \multicolumn{1}{c}{\textsc{B1}} & \multicolumn{1}{c}{\textsc{B3}} & \multicolumn{1}{c}{\textsc{B4}} & \multicolumn{1}{c}{score} \\ \midrule
\multirow{2}{*}{No-Reg} & \textsc{bs}            & -          & -              & -              & 16.100                          & \textbf{0.152}                  & {\ul \textbf{0.152}}            & {\ul \textbf{0.026}}            & {\ul \textbf{0.013}}                                 & {\ul \textbf{0.699}}               \\
                        & \textsc{con}           & 2          & -              & -              & 11.810                          & {\ul \textbf{0.156}}            & \textbf{0.152}                  & \textbf{0.020}                  & \textbf{0.009}                              & \textbf{0.337}                     \\ \midrule
\multirow{2}{*}{KLAR}   & \textsc{bs}            & -          & 0.1             & 0.4             & 16.332                          & 0.155                           & 0.152                           & 0.027                           & 0.015                                                      & {\ul \textbf{0.767}}               \\
                        & \textsc{con}           & 2          & 0.1             & 0.4             & 12.862                          & 0.156                           & 0.154                           & 0.021                           & 0.010                                                    & \textbf{0.707}                     \\ \midrule
\multirow{4}{*}{EAR}    & \textsc{con}           & 2          & 1              & -              & 13.091                          & {\ul \textbf{0.153}}            & 0.146                           & 0.016                           & 0.008                                                      & {\ul \textbf{0.523}}               \\
                        & \textsc{bs}            & -          & 0.01            & -              & 19.057                          & \textbf{0.152}                  & 0.149                           & 0.021                           & 0.010                                                       & \textbf{0.520}                     \\
                        & \textsc{con}           & 3          & 0.01            & -              & 11.204                          & 0.149                           & 0.150                           & 0.014                           & 0.006                                                      & 0.490                              \\
                        & \textsc{bs}            & -          & 1              & -              & 17.078                          & 0.145                           & 0.144                           & {\ul \textbf{0.026}}            & {\ul \textbf{0.015}}                                    & 0.490                \\ \bottomrule             
\end{tabular}
}
\caption{Results of the generation on the validation set.}
\label{tab:res_hyp}
\end{table*}

To select the decoding mechanisms and the regularization hyperparameters to be employed in our experiments, we fine-tune several models and generate 500 CNs by using the HS of the validation set as input. We then evaluate the generated CNs with the automatic metrics described in Section \ref{eval_metrics} and chose the combination of hyperparameters and decoding mechanisms which were giving the best results. 

For what regards the decoding mechanisms, we experimented with beam search \citep{li-etal-2016-deep, wiseman-etal-2017-challenges}, top-$k$ \cite{fan-etal-2018-hierarchical}, top-$p$ \citep{holtzman2019curious}, the combination of top-$k$ and top-$p$, and contrastive search \citep{su2022a}. For the decoding mechanisms hyperparameters we used the same as \citet{tekiroglu-etal-2022-using}, i.e. beam search with 5 beams and repetition penalty of 2; top-$k$ with $k = 40$, top-$p$ with $p = .92$, the combination of top-$p$ and top-$k$ with $k = 40$ and $p = .92$. For contrastive search, we used a $penalty\_alpha = 0.6$ and we tuned the $k$ parameter with a search space of ${2, 3, 5, 7, 9}$.

For EAR, we selected the $\alpha$ value from ${1, 0.1, 0.01}$; for KLAR we tuned both the $\alpha$ value (with the same search space as for EAR) and the percentage of attention to pose on relevant terms, selected among: {30\%, 40\%, 50\%, 60\%, 70\%}. For KLAR, we also tested two different configurations: one in which the special tokens \texttt{<HS>} and \texttt{<CN>} are considered together with the relevant terms receiving the special share of attention, and one configuration in which they do not.

In Table \ref{tab:res_hyp}, only the configurations achieving the best results for each regularization technique are showed. The column \textit{score} gives a general overview of the results. To obtain it, we first
normalized each metrics scores so that they fall into the 0-1 range, and then summed the mean of the normalised values. To calculate this score, we also considered the percentage of generation errors such as empty generations or generation of more than one \texttt{<HS>} or \texttt{<CN>} special token.

\begin{itemize}
    \item  No-Reg: the best results are achieved using beam search decoding strategy. However, this is also the configuration with the highest \textsc{RR} (16.1), which is 4.1 times the \textsc{RR} of the gold reference (3.89).
    For this reason, we choose to test also the second best scoring model, which uses the contrastive search decoding and achieves a \textsc{RR} of 11.8, without     a big impact on the the other metrics: this configuration still achieves the best or second best performance on ROUGE, BLEU and BERTscore.
    \item KLAR: the best configurations have $\alpha = 0.1$, share $= 0.4$ (i.e. 40\% of the attention is posed on the relevant terms) and using beam search and contrastive decoding. In both cases, \texttt{<HS>} and \texttt{<CN>} tags are not considered among the relevant terms receiving the special share of attention. It is once again a configuration achieving very high \textsc{RR} (16.33). Thus, we will also test the second best, which has a slightly worse performance on ROUGE, BLEU and BERTscore but a \textsc{RR} of 12.86.
    \item EAR: the best performing model has $\alpha = 1$ and uses contrastive search. The \textsc{RR} is high (13.09), but not the highest (which is 19.06). As second configuration, we choose to test the same setup but with beam search, in order to have comparable results. Also, this configuration is still the one achieving the best results on BLEU-3 and BLEU-4.

\end{itemize}

\end{document}